\pdfoutput=1

\documentclass[11pt]{article}

\usepackage[preprint]{acl}

\usepackage{times}
\usepackage{latexsym}
\usepackage[T1]{fontenc}
\usepackage[utf8]{inputenc}
\usepackage{microtype}
\usepackage{inconsolata}
\usepackage{graphicx}
\usepackage{booktabs}
\usepackage{xcolor,pifont}
\usepackage{fontawesome}

\newcommand \blfootnote[1]{
    \begingroup
        \renewcommand
        \thefootnote{}\footnote{#1}
        \addtocounter{footnote}{-1}
        \vspace{-1ex}
    \endgroup
}

\title{Pre-trained Language Models Do Not Help Auto-regressive \\ Text-to-Image Generation}

\author{
\textsuperscript{\faApple}Yuhui Zhang\textsuperscript{1},
\textsuperscript{\faApple}Brandon McKinzie\textsuperscript{2},
Zhe Gan\textsuperscript{3},
Vaishaal Shankar\textsuperscript{3},
Alexander Toshev\textsuperscript{3} \\
\textsuperscript{1}Stanford University,
\textsuperscript{2}OpenAI,
\textsuperscript{3}Apple ML Research \\
\small{\textbf{Correspondence:} \href{mailto:yuhuiz@stanford.edu}{yuhuiz@stanford.edu}}
\vspace{-1em}
}

\begin{document}
\maketitle
\begin{abstract}
Recent advances in image tokenizers, such as VQ-VAE, have enabled text-to-image generation using auto-regressive methods, similar to language modeling. However, these methods have yet to leverage pre-trained language models, despite their adaptability to various downstream tasks. In this work, we explore this gap by adapting a pre-trained language model for auto-regressive text-to-image generation, and find that pre-trained language models offer limited help. We provide a two-fold explanation by analyzing tokens from each modality. First, we demonstrate that image tokens possess significantly different semantics compared to text tokens, rendering pre-trained language models no more effective in modeling them than randomly initialized ones. Second, the text tokens in the image-text datasets are too simple compared to normal language model pre-training data, which causes the catastrophic degradation of language models' capability.~\blfootnote{\textsuperscript{\faApple}Work done while at Apple ML Research.}
\end{abstract}

\section{Introduction}

\begin{figure*}[htbp]
    \centering
    \includegraphics[width=\linewidth]{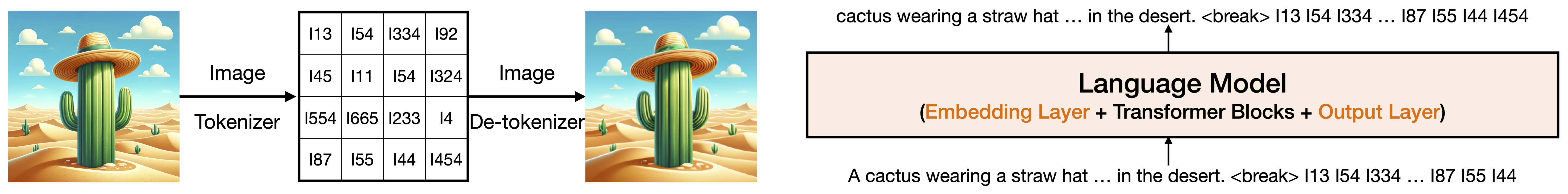}
    \vspace{-6mm}
    \caption{
    \emph{Adapting language models for auto-regressive text-to-image generation.} \textit{(Left)} An image is fed into an image tokenizer (MoVQGAN~\citep{zheng2022movq}) and converted to a grid of discrete tokens, and it can be well-reconstructed with these image tokens. \textit{(Right)} As images are converted to tokens similar to text tokens, we can enable language models to generate images by adapting its embedding layer and output layer. }
    \label{fig:method}
    \vspace{-3mm}
\end{figure*}

Recent works in text-to-image generation primarily employ two kinds of methods: diffusion models~\citep{ramesh2022hierarchical,saharia2022photorealistic,rombach2022high} and auto-regressive models~\citep{dalle,yu2022scaling}. The latter is facilitated by ``image tokenizers'', such as VQ-VAE~\citep{van2017neural,razavi2019generating} and VQ-GAN~\citep{esser2021taming,yu2021vector}, which transform an image into a sequence of discrete tokens, similar to text tokens (Figure~\ref{fig:method} Left). Consequently, image and text tokens can be jointly modeled using auto-regressive algorithms like the Transformer~\citep{vaswani2017attention} (Figure~\ref{fig:method} Right). 

\begin{figure}[t!]
    \centering
    \includegraphics[width=\linewidth]{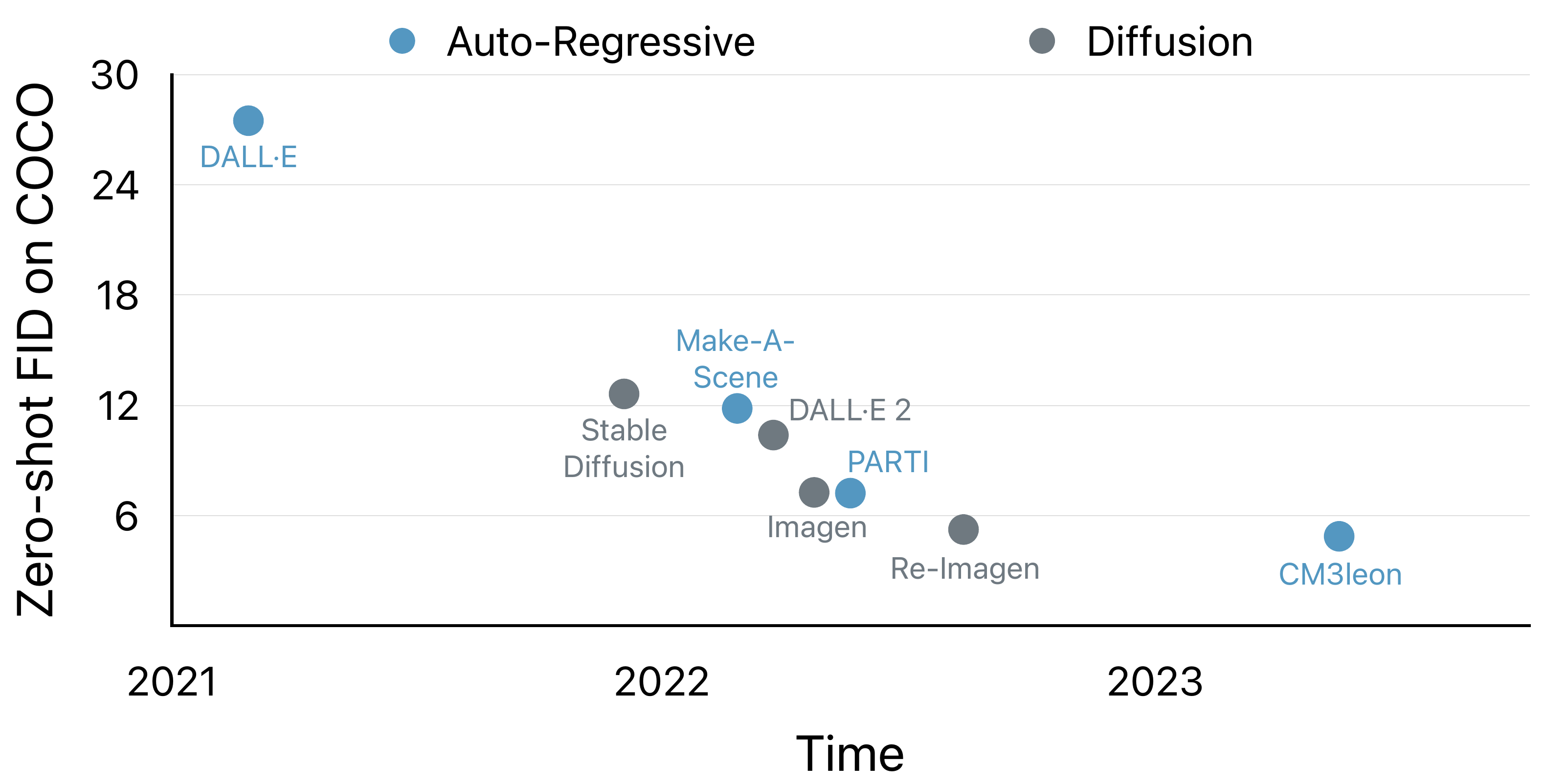}
    \vspace{-8mm}
    \caption{\emph{Auto-regressive and diffusion based models achieve similar performances on text-to-image generation.} However, while all the diffusion models leverage pre-trained language models, all the auto-regressive models do not. }
    \label{fig:related-works}
    \vspace{-3mm}
\end{figure}

The superiority of diffusion-based models when compared with auto-regressive-based methods for text-to-image generation still remains unclear. OpenAI's pioneering work, DALL-E~\cite{dalle}, showcased the potential of auto-regressive methods in this domain. Yet, its successor, DALL-E 2~\cite{ramesh2022hierarchical}, transitioned to a diffusion-based architecture and achieved enhanced image generation quality. Later, Google released Imagen~\cite{saharia2022photorealistic} (diffusion-based) and Parti~\cite{yu2022scaling} (auto-regressive-based) at the same time and demonstrated their comparable generation quality. Similarly, the retrieval-augmented methods, Re-Imagen~\cite{reimagen} (diffusion-based) and CM3leon~\cite{cm3leon} (auto-regressive-based), display similar performance in text-to-image generation tasks. A comparison based on zero-shot FID~\cite{heusel2017gans} on the COCO dataset~\cite{coco} can be found in Figure~\ref{fig:related-works}.

While these two approaches achieve similar performance, it is intriguing that \emph{diffusion-based models consistently utilize pre-trained text encoders, whereas their auto-regressive counterparts generally do not}. For instance, Imagen~\cite{saharia2022photorealistic} (diffusion-based) reports that employing a stronger pre-trained text encoder, specifically T5~\citep{raffel2020exploring}, yields substantial improvements to using CLIP~\citep{radford2021learning}. Furthermore, they observe that scaling up the T5 text encoder leads to more pronounced improvements than scaling up the diffusion models. Conversely, Parti~\cite{yu2022scaling} (auto-regressive-based) shows that using a pre-trained text encoder does not necessarily improve image quality in its Appendix. However, Parti employs an encoder-decoder architecture and uses BERT~\citep{devlin2018bert}, a relatively inferior text encoder, to initialize the encoder only. It remains unclear whether a decoder-only approach would benefit from recent advances in large language models (LLMs), given the clear similarity between language modeling and auto-regressive text-to-image generation.

In this work, we explore the potential of pre-trained LLMs for auto-regressive text-to-image generation. To enable the model to process both text and image tokens, we expand the size of the embedding and output layers by incorporating an image vocabulary from the image tokenizer. We initialize these added weights either randomly or using a novel contrastive alignment (elaborated later in Section~\ref{sec:alignment}), while the remaining weights are directly copied from the original models. Subsequently, we fine-tune the model on image-caption datasets, as depicted in Figure~\ref{fig:method} Right. 

\begin{figure*}[t!]
    \centering
    \includegraphics[width=0.75\linewidth]{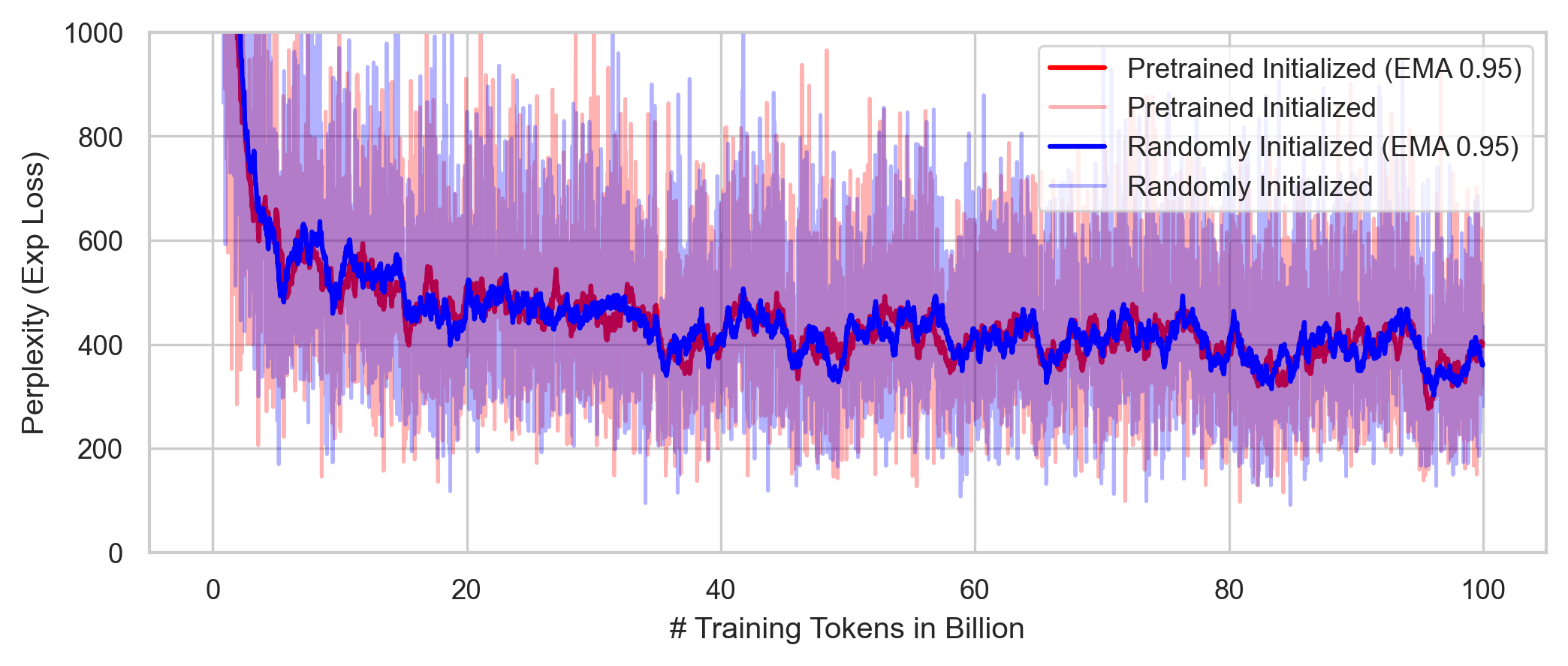}
    \vspace{-3mm}
    \caption{\emph{Pre-trained language models do not help auto-regressive text-to-image generation.} Models are trained on the HQITP-134M image-caption dataset with 64 A100 80GB GPUs using batch size 1M tokens. EMA is Exponential Moving Average. }
    \vspace{-2mm}
    \label{fig:main-result}
\end{figure*}

Surprisingly, the results show that pre-trained language models achieve the same loss and image generation quality as the model that is entirely randomly initialized and trained from scratch (Figure~\ref{fig:main-result}). Furthermore, we observe a catastrophic deterioration in the model's text capabilities, such as world knowledge or in-context learning, after only minimal steps of fine-tuning (Table~\ref{tab:forgetting}). 

To understand this phenomenon, we break down the cross-entropy loss on image and text tokens, and find that 1) the loss on image tokens is the same between the pre-trained and randomly initialized model, and 2) the loss on text tokens of the pre-trained model is significantly lower at the beginning compared to the randomly initialized models, but the gap soon disappears after training (Figure~\ref{fig:mixed-token}). 

The first finding of the loss on the image tokens is particularly interesting. We hypothesize that image tokens obtained from image tokenizers might either lack semantics or possess significantly different semantics compared to text tokens, which renders language pre-training not transferable to the image modeling task. To verify this hypothesis, we conduct unconditional image generation experiments by training the model on image tokens only. Our results show that 1) the pre-trained model achieves the same loss as the randomly initialized model, and 2) freezing any part of the pre-trained model results in a loss degradation (Figure~\ref{fig:image-token-only}). These indicate that optimal weights for language and image modeling are fundamentally different, making language pre-training not transferable to image modeling. 

In summary, we share our experimental findings about pre-trained language models do not help auto-regressive text-to-image generation, and offer an explanation: 1) the intrinsic differences between image and text tokens make language pre-training ineffective for the image token modeling, and 2) the disproportionate ratio between image and text tokens (usually 30:1 for image-caption datasets) minimizes the impact of loss on text tokens and leads to catastrophic forgetting. 

\section{Pre-trained Language Models Do Not Help Text-to-Image Generation}

\subsection{Experimental Setup}

\paragraph{Language model.} We use the publicly available \texttt{open\_lm} codebase and its \texttt{open\_lm-1b} model for our experiments~\cite{openlm}. This language model contains $\sim$1B parameters and is trained on 1.6T tokens on a mix of RedPajama~\cite{redpajama}, Pile~\cite{pile}, S2ORC~\cite{s2orc}, The Pile of Law~\cite{pile-of-law}, Deepmind Math~\cite{deepmind-math}, and RealNews~\cite{realnews}. It achieves better or comparable performance compared to models with similar size such as OPT-1.3B~\cite{opt}, Pythia-1B~\cite{pythia}, Neox-1.3B~\cite{neox}, OPT-IML-1.3B~\cite{opt-iml} on an average of 11 tasks such as HellaSwag~\cite{hellaswag} and MMLU~\cite{mmlu}. More details can be found in the \texttt{open\_lm} repository~\cite{openlm}.

\paragraph{Image tokenizer.} We use SBER-MoVQGAN~\cite{zheng2022movq} as the image tokenizer, which is the current state-of-the-art publicly available image tokenizer that achieves 0.686 FID on Imagenet image reconstruction. Given an image with 256 $\times$ 256 resolution, it converts an image to 1,024 tokens with a vocabulary size of 16,384. Figure~\ref{fig:method} (Left) shows a real reconstruction example from this tokenizer.  

\paragraph{Dataset.} For multi-modal training, we use an internal dataset referred to as High Quality Image-Text Pairs (HQITP)~\cite{ranasinghe2023}, which contains 134M high-quality image-caption pairs. 
The primary sources of image-caption pairs in HQITP are from the web, similar to the commonly used image-caption datasets such as Conceptual Captions (CC)~\cite{changpinyo2021cc12m}. 
We chose HQITP because it is larger, has higher quality, and includes a broader range of concepts and objects, thus validating our conclusions on a larger scale.
Previous works leveraging HQITP have shown that conclusions transfer well between HQITP and CC~\cite{ranasinghe2023perceptual}.

We pre-process the dataset before training. Each image is center-cropped to 256 $\times$ 256 and converted to 1,024 tokens. Each caption is tokenized with NeoX tokenizer with an average of 30 tokens. We add six special tokens corresponding to the beginning and end of document, text segment, and image, respectively. This results in input sequences of the form ``<doc> <text> ...text tokens... </text> <image> ...image tokens... </image> </doc>'', and pad them into 1,152 tokens with the special <pad> token. 

\paragraph{Training setups.} Models are trained with 100B tokens using 64 A100 80GB GPUs with batch size 1M tokens. We use the AdamW~\cite{adamw} optimizer with a cosine learning rate schedule with 2K warm-up steps and a peak learning rate of 0.0003. This mimics the settings reported in~\cite{multimodal-scaling-laws}. We also tried different hyperparameters, such as learning rates from 0.00005 to 0.0003 and batch size from 0.5M to 2M tokens, and found no significant influences on the conclusions.

\subsection{Results}

\begin{figure*}[t!]
    \centering
    \includegraphics[width=0.45\linewidth]{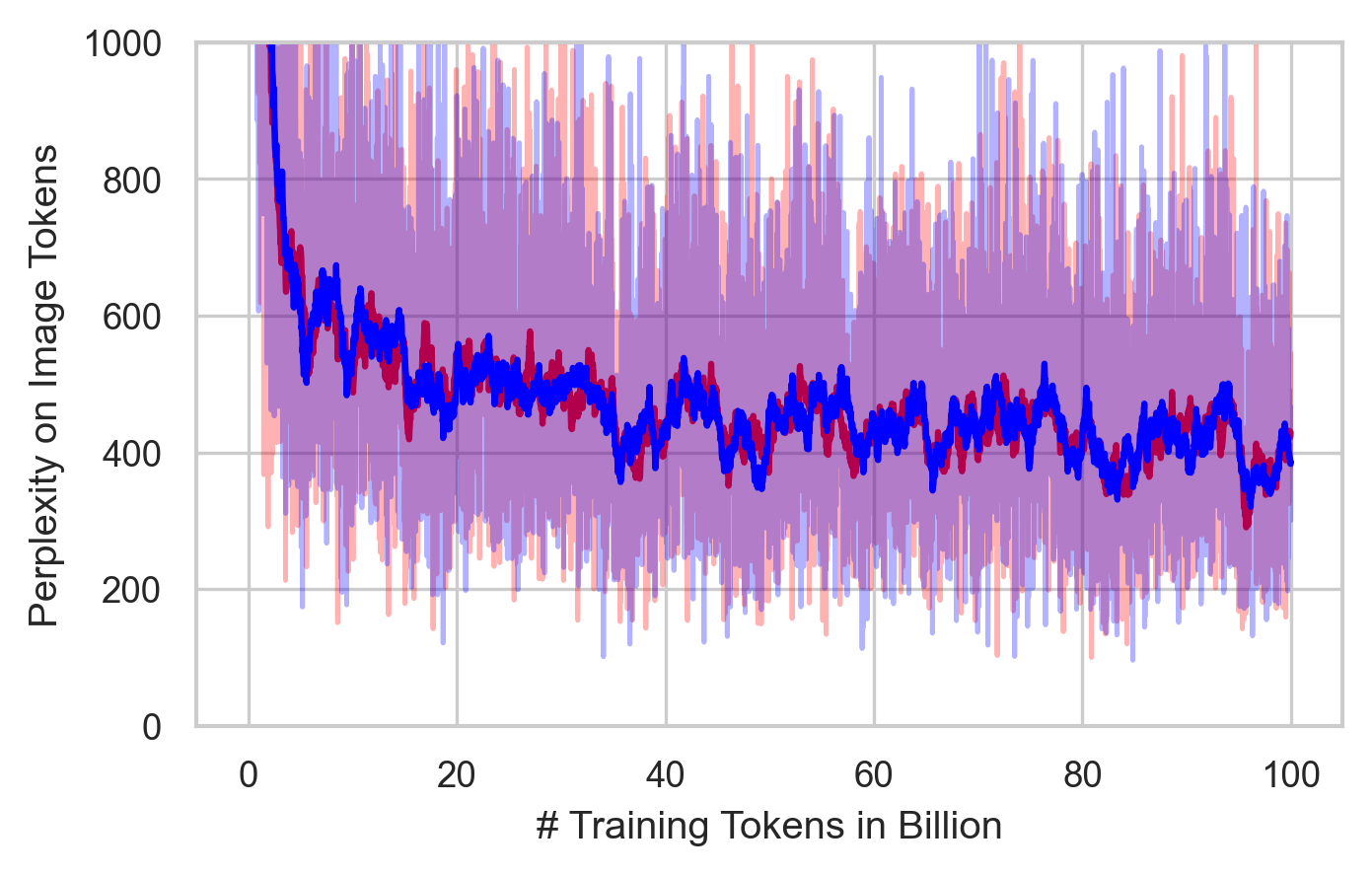}
    \includegraphics[width=0.45\linewidth]{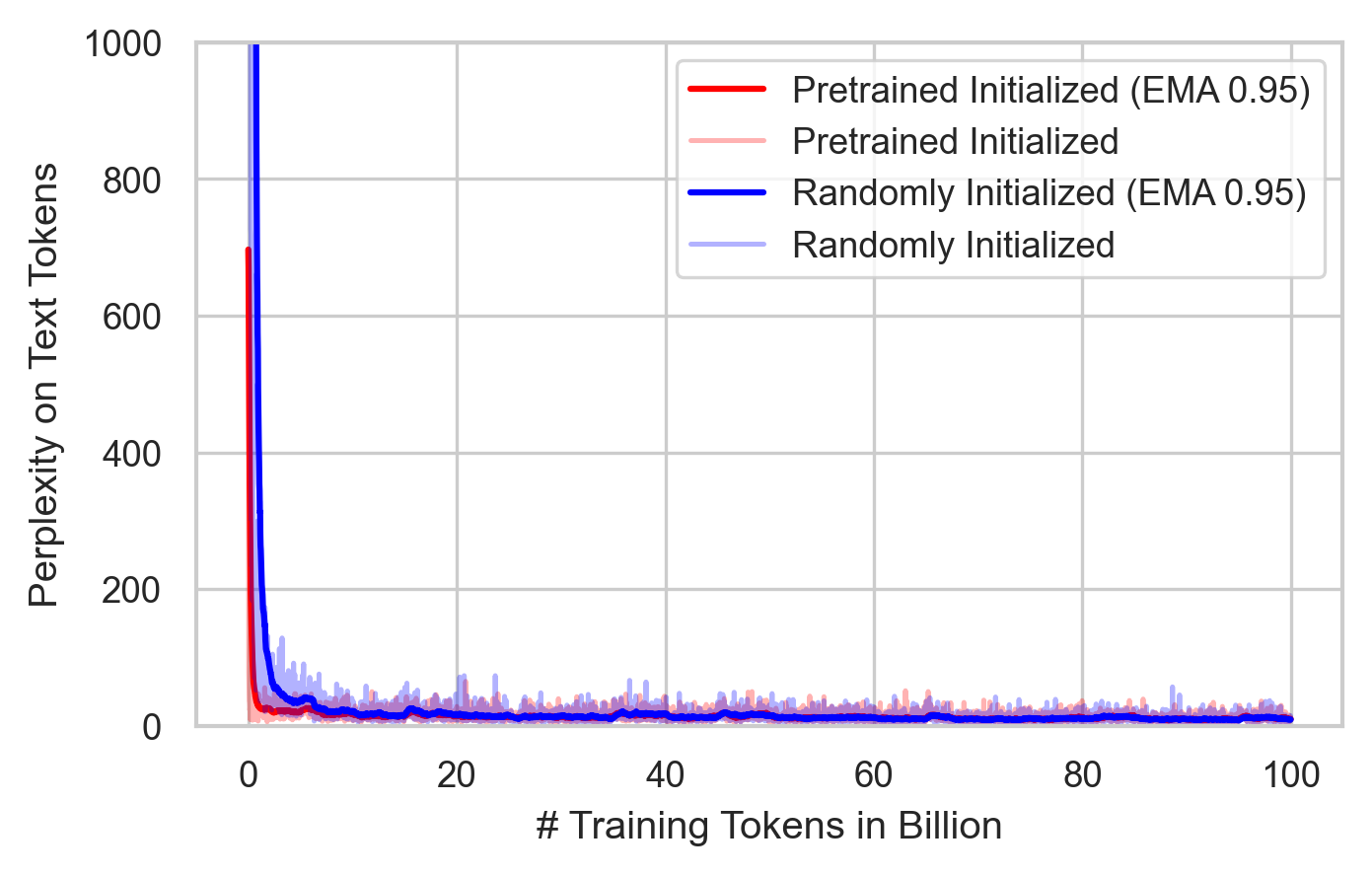}
    \vspace{-3mm}
    \caption{\emph{Break-down loss on image and text tokens.} Models are trained on the HQITP-134M image-caption dataset with 64 A100 80GB GPUs using batch size 1M tokens. }
    \label{fig:mixed-token}
    \vspace{-3mm}
\end{figure*}

In Figure~\ref{fig:main-result}, we present the perplexity (exponential of loss) during training for both the pre-trained and randomly initialized models. Intriguingly, across the entire 100B token training regimen, the loss of the pre-trained model aligns closely with that of the randomly initialized one. Beyond this, a sharp decline in text capabilities of the pre-trained model is observed after training on 5B tokens, as illustrated in Table~\ref{tab:forgetting}. At this point, both the model's world knowledge and its in-context learning ability are entirely diminished.

To delve deeper into this phenomenon, we separate the cross-entropy loss into two components: text tokens and image tokens, displayed separately in Figure \ref{fig:mixed-token}. As anticipated, the pre-trained model begins with a significantly lower text loss in comparison to its randomly initialized counterpart. Yet, due to the overwhelming image-text token ratio (30:1), this initial advantage is obscured in the aggregate loss. Furthermore, any benefit the pre-trained model offers in text loss diminishes soon during training. 
In contrast, for image tokens, there is no difference between the pre-trained and randomly initialized models. We hypothesize that the inability of effectively transferring a pre-trained language model to image token modeling is caused by the distinction between image and text tokens.

Moreover, loss on text tokens is substantially lower than image tokens, and even lower than typical language models trained on text-only data. This is because texts in image-caption datasets such as HQITP are less complex than those in standard text-only pre-training corpora, which also explains the catastrophic degradation of the model's text capability. 

\begin{table}[t!]
\small
    \centering
    \begin{tabular}{p{0.47\linewidth}|p{0.47\linewidth}}
    \toprule
    Original Completion & Completion after Training 5B Tokens \\
    \midrule
      Simply put, the theory of relativity states that \textbf{the speed of light is the same for all observers, regardless of their location in the universe.}   & Simply put, the theory of relativity states that \textbf{iles must be able to see the invisible.} \\
    \midrule
      Translate English to French: & Translate English to French: \\
sea otter => loutre de mer &   sea otter => loutre de mer \\
peppermint => menthe poivrée & peppermint => menthe poivrée \\
plush girafe => girafe peluche & plush girafe => girafe peluche \\
cheese => \textbf{fromage}  &    cheese => \textbf{I love cheese} \\
    \bottomrule
    \end{tabular}
    \vspace{-2mm}
    \caption{\emph{Concrete examples of forgetting.} We observe a severe deterioration of the model's language capability, such as knowledge and in-context learning, after a small amount of training. Model completions are bolded. }
    \label{tab:forgetting}
    \vspace{-3mm}
\end{table}
 
\begin{figure}[t!]
    \centering
    \includegraphics[width=0.85\linewidth]{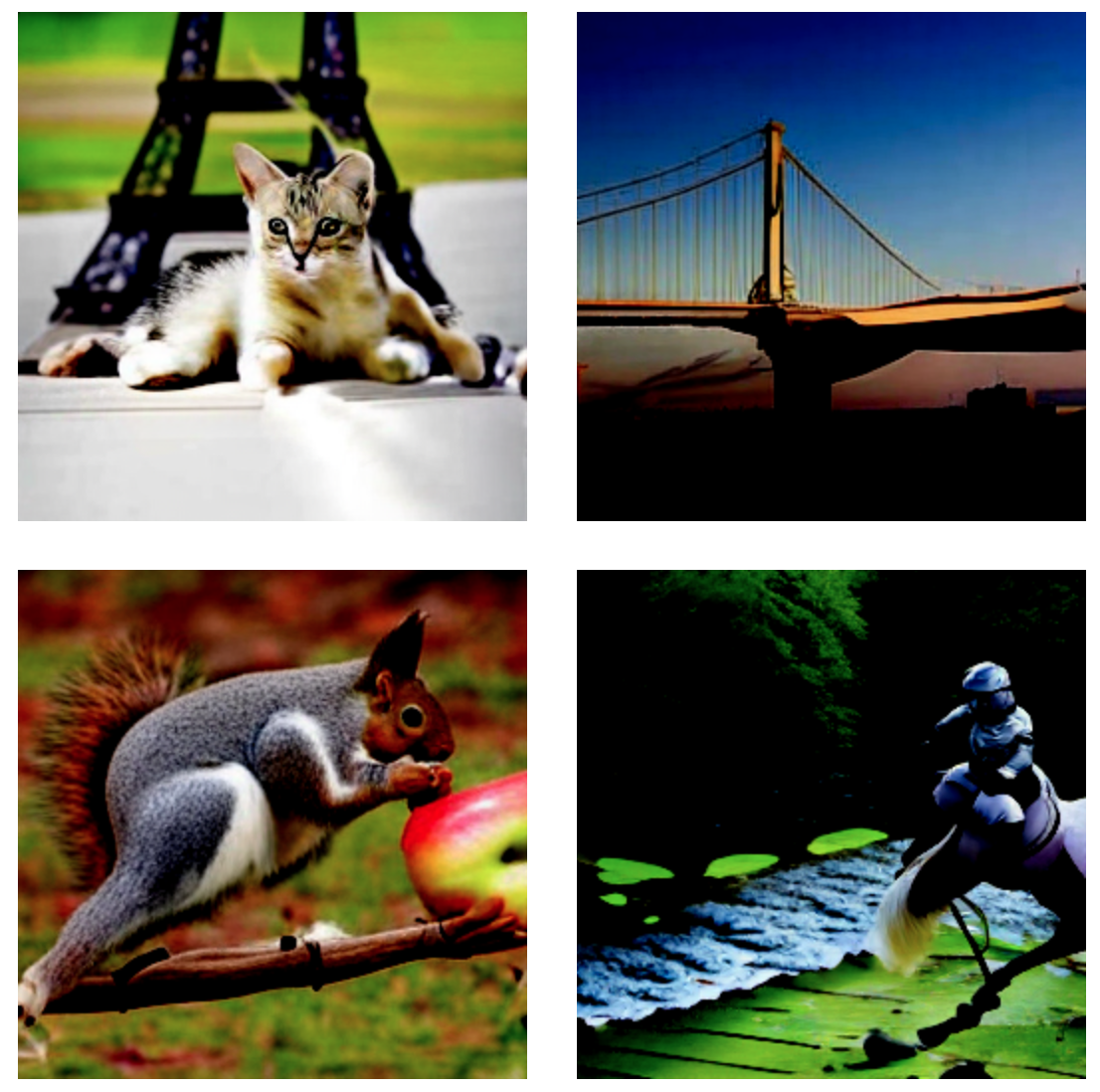}
    \vspace{-3mm}
    \caption{\emph{Examples of generated images.} We achieve 12.21 FID on MS-COCO at the end of training.}
    \label{fig:generation}
    \vspace{-3mm}
\end{figure}

We use perplexity as our main evaluation metric for its ability to provide finer-grained insights into training dynamics, which is essential for our conclusion that pre-trained language models do not enhance auto-regressive text-to-image generation. Unlike time-consuming metrics like FID (Fréchet Inception Distance)~\cite{heusel2017gans}, perplexity is computationally inexpensive and allows us to compare models at nearly every training step. Our results show that perplexity on image tokens is nearly identical for both pre-trained and randomly initialized models, supporting our claim. Additionally, FID scores at the end of training on MS-COCO further validate this, with both models showing nearly identical performance (12.21 for pre-trained language models vs. 12.27 for randomly-initialized language models), demonstrating that pre-training offers no significant advantage in this setting. FID scores are slightly below DALL-E 2~\cite{ramesh2022hierarchical}, due to training on only 100B tokens; continued training enhances quality. We provide some generation examples in Figure~\ref{fig:generation}.

\section{Image Tokens Are Drastically Different From Text Tokens}

Why there is no difference between the loss of pre-trained and randomly initialized models on the image tokens? We hypothesize image tokens are significantly different from text tokens, for example, they lack semantics or have drastically different semantics compared to text tokens, which makes the pre-trained language model not transferable to image token modeling. Our unconditional image generation and image-token alignment experiments verify this hypothesis.

\subsection{Unconditional Image Generation}

\begin{figure}[t!]
    \centering
    \includegraphics[width=\linewidth]{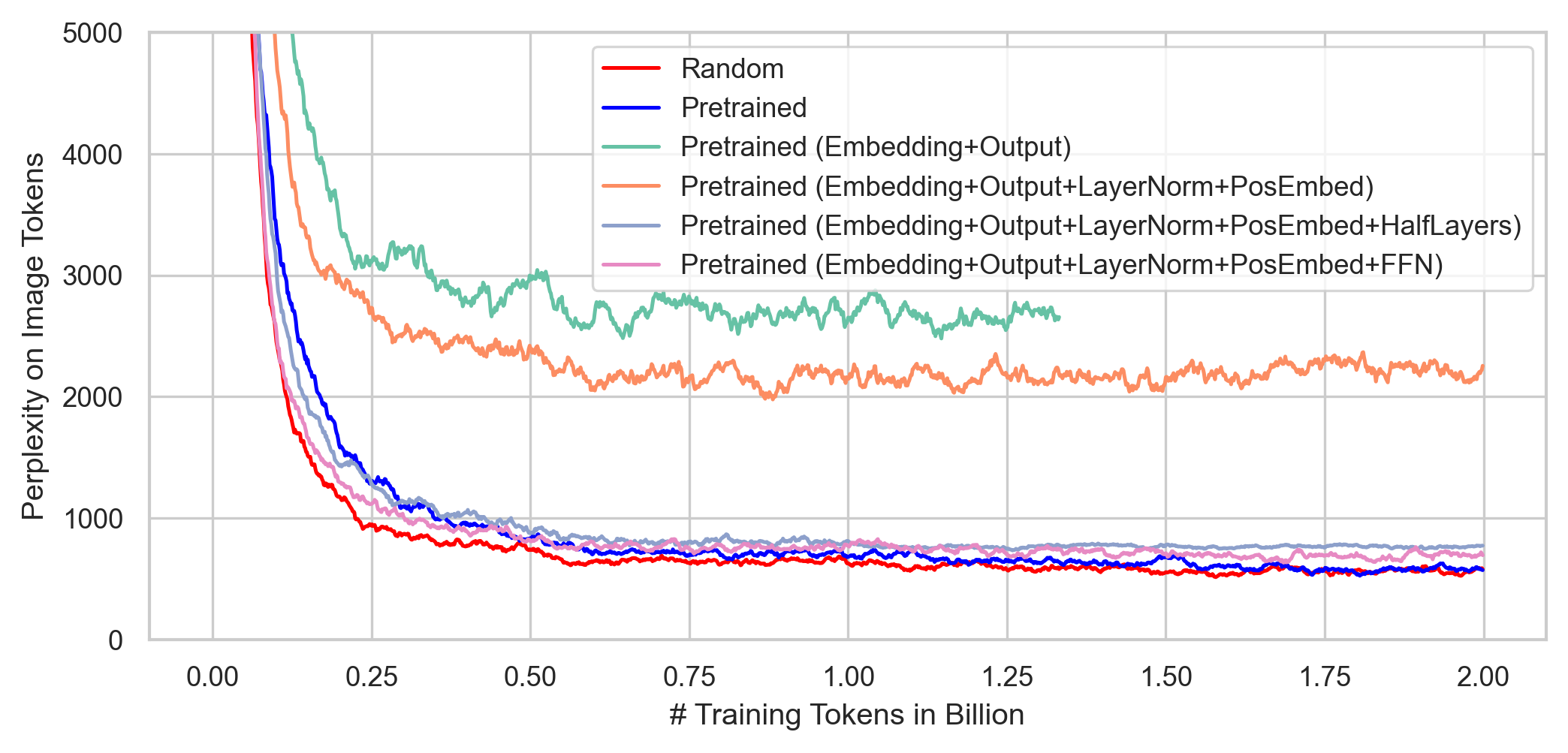}
    \vspace{-8mm}
    \caption{\emph{Pre-trained language models do not help to model image tokens.} Models are trained only on the HQITP dataset's image tokens without any text tokens. We also compare the full fine-tuning with electively fine-tuning components of the pre-trained models (shown in parenthesis). EMA 0.95 is applied to the plot.}
    \label{fig:image-token-only}
\end{figure}

To assess if pre-trained language models benefit image tokens, we perform unconditional image generation experiments. Unlike the text-to-image generation setup, we removed all text tokens, leaving only the image tokens. This approach rigorously examines if image tokens benefit from pre-trained language models. As shown in Figure~\ref{fig:image-token-only}, pre-trained language models yield the same loss as models initialized randomly.

Additionally, we selectively tune components of the pre-trained models: 1) only the embedding and output layer; 2) 1 plus layer norm and positional embedding; and 3) 2 plus the first half of layers; 4) 2 plus the feed-forward layers (FFN). Figure~\ref{fig:image-token-only} presents these loss metrics. The findings reveal that none of these configurations achieves as low a loss as a fully tunable model. This underscores the divergence in optimal weights for modeling text and image tokens, suggesting that any part of the text-trained weights is sub-optimal to transfer to image tokens.

\subsection{Image-Text Token Contrastive Alignment}
\label{sec:alignment}

To understand whether image tokens have similar semantics as text tokens, we aligned image tokens with text tokens using a contrastive approach, inspired by methods like CLIP~\cite{radford2021learning}. Given an image, we tokenize it into 1024 tokens and compute its bag-of-words image embeddings as its representation. Similarly, we tokenize the corresponding caption and compute its bag-of-words text embeddings. The text embeddings are initialized from a pre-trained language model while the image embeddings are randomly initialized. For a batch of \(N=1024\) image-caption pairs, the contrastive objective from CLIP is employed to maximize the cosine similarity between matched image-caption $l_2$-normalized representations and to minimize the similarity for non-matching pairs. Only the image embeddings are updated during training.

\begin{figure}[t!]
    \centering
    \includegraphics[width=\linewidth]{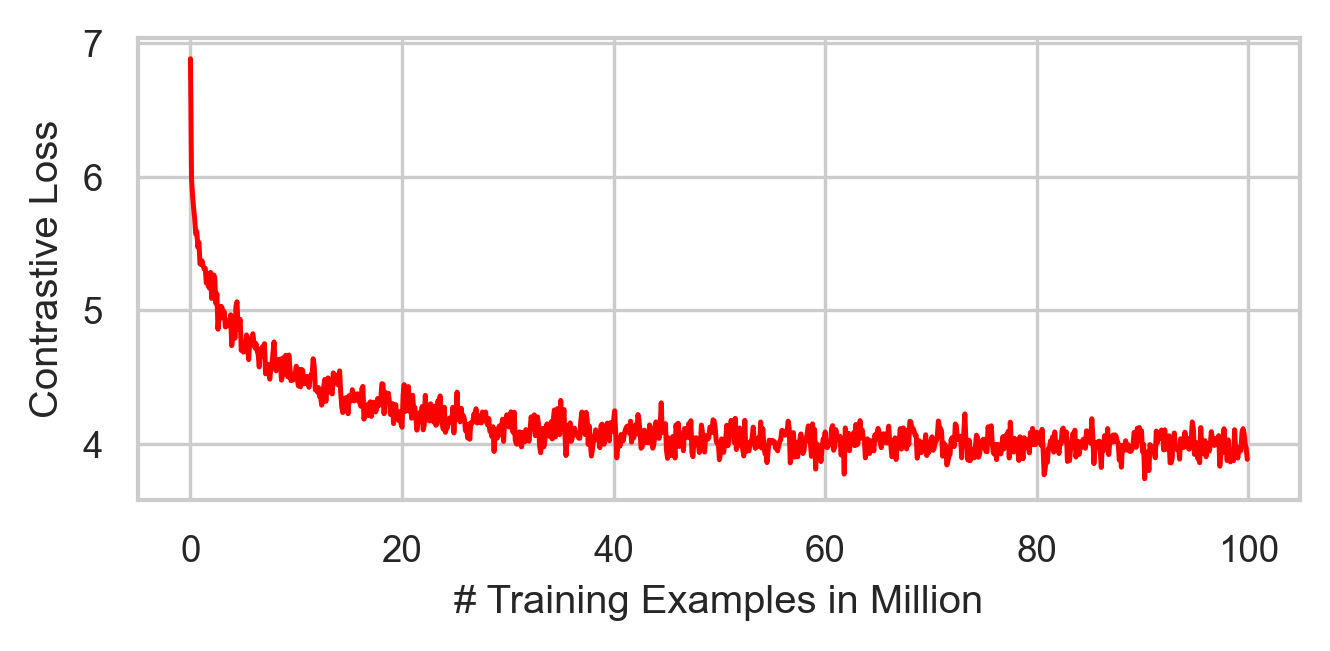}
    \includegraphics[width=\linewidth]{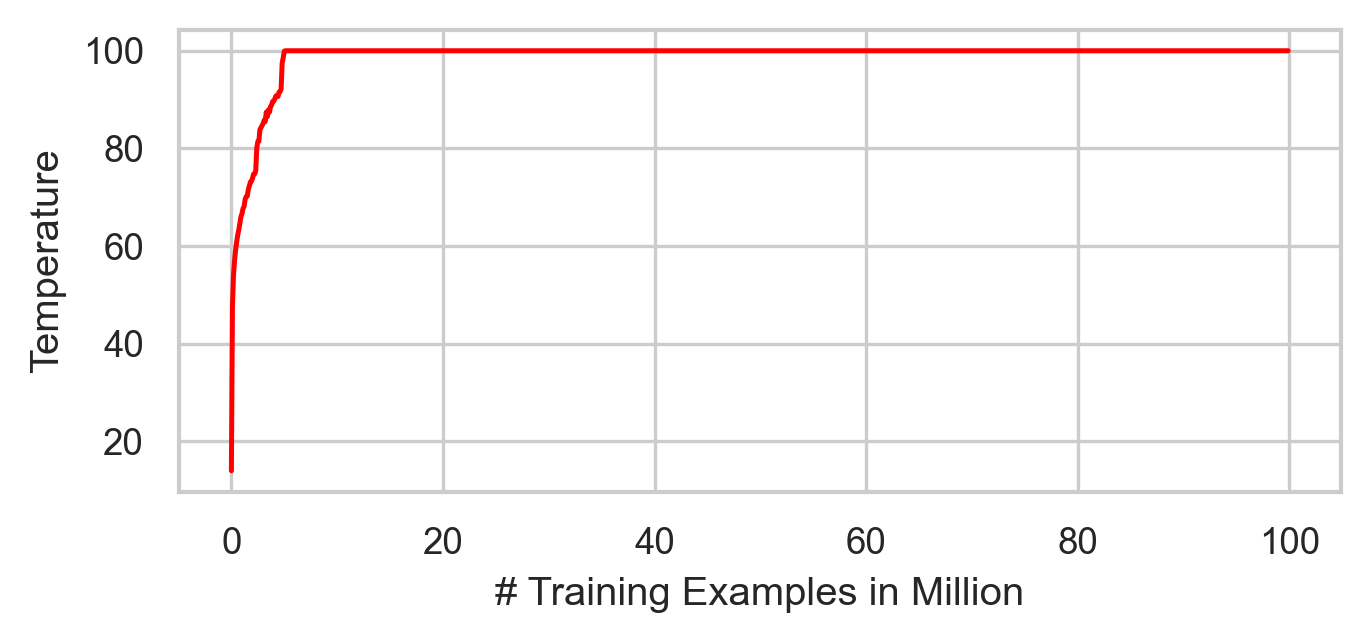}
    \vspace{-6mm}
    \caption{\emph{Image-text token contrastive alignment.} \textit{(Top)} The contrastive loss plateaus quickly, indicating a difficulty in aligning text and image tokens directly at a bag-of-words level. \textit{(Bottom)} The learnable temperature in the contrastive loss during training for reference.  }
    \label{fig:token-alignment}
    \vspace{-3mm}
\end{figure}

In Figure~\ref{fig:token-alignment}, we illustrate that the contrastive loss plateaus quickly, indicating a difficulty in aligning text and image tokens directly at a bag-of-words level. Indeed, after training, when querying the closest text tokens for any image token, we observe that they predominantly align with noisy, semantically void text tokens. Furthermore, when we use the trained image embeddings as initialization for text-to-image generation, as opposed to random initialization, there is no discernible improvement.

\section{Conclusion}

This study highlights the difficulty of naively adapting a text-only language model to handle multi-modal contents, such as texts and images. Given the challenge of the disparities between image tokens and text tokens, a valuable avenue for future experiments is to employ tokenizers that align semantically with text tokens, such as SEED~\cite{seed} or SPAE~\cite{spae}.

\section*{Limitations}

Our study has some limitations. First, the results are based on the VQGAN image tokenizer, which does not align semantics between image tokens and text tokens. Tokenizers that semantically align image tokens with text tokens might yield different outcomes. Second, we observed severe degradation in language model capabilities during fine-tuning, suggesting that exploring methods to avoid catastrophic forgetting could be a promising future research direction. Additionally, our experiments used internal image-caption datasets and required extensive computational resources, which might limit the reproducibility of exact numbers. Despite these limitations, our findings remain useful and transferable and provide valuable information for future research.

\bibliography{latex/main}

\end{document}